\documentclass[a4paper,nobib]{tufte-handout}
\input{bibstyle}

\usepackage{graphicx}
\usepackage{booktabs}
\usepackage{csquotes}
\usepackage{xspace}
\usepackage{amsmath}
\usepackage{amssymb}
\usepackage{mathtools}
\usepackage{amsthm}
\usepackage{thmtools, thm-restate}
\usepackage{centernot}

\theoremstyle{plain}

\theoremstyle{definition}

\theoremstyle{remark}

\usepackage{hyperref}
\hypersetup{
    colorlinks,
    linkcolor={red!50!black},
    citecolor={blue!50!black},
    urlcolor={blue!50!black}
}

\usepackage{fancyhdr}

\makeatletter
\fancypagestyle{tufte-fancy}{
  \fancyhf{}
  \fancyhead[RE,RO]{\textsc{a primer on causal and statistical dataset biases}\quad\thepage}

}
\makeatother
\newcommand{\iid}{\textsc{iid}\xspace}
\newcommand{\erm}{\textsc{erm}\xspace}
\newcommand{\scm}{\textsc{scm}\xspace}
\newcommand{\frl}{\textsc{frl}\xspace}
\newcommand{\auc}{\textsc{auc}\xspace}

\usepackage[capitalize]{cleveref}
\crefname{assumption}{Assumption}{Assumptions}

\crefformat{chapter}{\S#2#1#3}
\crefformat{section}{\S#2#1#3}
\crefformat{appendix}{\S#2#1#3}
\crefformat{subsection}{\S#2#1#3}
\crefformat{subsubsection}{\S#2#1#3}
\crefrangeformat{section}{\S\S#3#1#4 to~#5#2#6}
\crefmultiformat{section}{\S\S#2#1#3}{ and~#2#1#3}{, #2#1#3}{ and~#2#1#3}

\usepackage{tikz}
\usepackage{array}
\usepackage{multirow,makecell}
\usepackage{collcell}
\definecolor{lred}{RGB}{255,179,186}
\definecolor{lred}{RGB}{255, 198, 194}
\definecolor{lred}{RGB}{254, 197, 187}
\definecolor{lyellow}{RGB}{255,211,93}
\definecolor{lyellow}{RGB}{167, 199, 231}
\definecolor{lyellow}{RGB}{255, 223, 200}
\definecolor{lyellow}{RGB}{250, 247, 240}
\definecolor{lgreen}{RGB}{169,215,151}
\definecolor{lgreen}{RGB}{195, 224, 221}

\newcommand\ColorBox[2]{%
  \tikz[baseline=(X.base)]\node[draw,fill=#1,anchor=base,minimum width=3.4cm,minimum height=3.4cm](X){\centering #2};%
}

\newcommand\MyBox[3]{%
  \ColorBox{#3}{\parbox{3.2cm}{\centering #1\\#2}}%
}

\newcommand{\ra}[1]{\renewcommand{\arraystretch}{#1}}

\title{A Primer on Causal and Statistical Dataset\\ Biases for Fair and Robust Image Analysis}
\author{Charles Jones and Ben Glocker}

\date{Imperial College London}

\begin{document}

\maketitle

\marginnote{Excerpt from C.J.'s PhD thesis. Winner of the G-Research PhD prize 2025.}

\begin{abstract}

\noindent Machine learning methods often fail when deployed in the real world. Worse still, they fail in high-stakes situations and across socially sensitive lines. These issues have a chilling effect on the adoption of machine learning methods in settings such as medical diagnosis, where they are arguably best-placed to provide benefits if safely deployed. In this primer, we introduce the causal and statistical structures which induce failure in machine learning methods for image analysis. We highlight two previously overlooked problems, which we call the \textit{no fair lunch} problem and the \textit{subgroup separability} problem. We elucidate why today's fair representation learning methods fail to adequately solve them and propose potential paths forward for the field.

\end{abstract}

\section{Introduction}

If we wish to deploy deep predictive models in high-stakes settings, such as medical diagnosis, we must understand and mitigate performance disparities across population subgroups \citep{seyyed-kalantariUnderdiagnosisBiasArtificial2021}. Despite considerable effort in developing methods for debiasing representations of deep models, little progress has been made towards understanding the validity of such methods for real-world deployment. Proposed methods often achieve state-of-the-art results on one benchmark, only to be beaten by conventional empirical risk minimisation \citep[\erm;][]{vapnikOverviewStatisticalLearning1999} on more comprehensive evaluations \citep{zietlowLevelingComputerVision2022,zongMEDFAIRBenchmarkingFairness2023}. Further analyses have shown a concerning `levelling down' effect \citep{mittelstadtUnfairnessFairMachine2023}, indicating that today's group fairness methods may even cause harm if deployed in the real world.

\marginnote{
\textbf{Acronyms and notation:}
\begin{itemize}
    \itemsep0em 
    \item[\scm] Structural causal model.
    \item[\erm] Empirical risk minimisation.
    \item[\frl] Fair representation learning.
    \item[\iid] Independent and identically distributed.
    \item[\auc] Area under receiver operating characteristic curve.
\end{itemize}

\begin{itemize}
    \itemsep0em 
    \item[$X$] Random variable.
    \item[$x$] Scalar realisation of $X$.
    \item[$\mathfrak{C}$] Structural causal model.
    \item[$p^\mathfrak{C}$] Density function induced by $\mathfrak{C}$.
    \item[$\mathbb{S}(X)$] Statistical support of $X$.
    \item[$\mathbb{1}_{\mathcal{X}}(x)$] Indicator (= 1 iff $x \in \mathcal{X}$).
    \item[$X \perp^\mathfrak{C} Y$] Statistical independence in $\mathfrak{C}$.
    \item[$I^\mathfrak{C}(X;Y)$] Mutual information in $\mathfrak{C}$.
\end{itemize}
}

In this primer, we show how many of these issues stem from an inadequate understanding of the underlying causal and statistical structures present in data. We analyse implicit assumptions harming the validity of machine learning methods and applications, and we produce theoretical and empirical results that explain apparent contradictions in the existing literature. Our work represents a step towards the safe deployment of machine learning methods in high-stakes settings. Our contributions are:

\begin{itemize}
    \itemsep0em 
    \item[\cref{sec:no-fair}] We demonstrate how a causal formulation of dataset bias provides a unifying perspective on fairness and distribution shift.
    \item[\cref{sec:sub-sep}] We introduce the statistical property of \textit{subgroup separability} and highlight how it affects the training of machine learning models.
    \item[\cref{sec:rethinking-frl}] We combine our causal and statistical analysis to understand the circumstances under which the class of \textit{fair representation learning} methods fails to adequately mitigate dataset bias.
\end{itemize}

\newpage
\section{A causal perspective on dataset bias} \label{sec:no-fair}

\marginnote[10em]{
This section is adapted from \citet{jonesCausalPerspectiveDataset2024} (\textit{Nature Machine Intelligence}). We include the causal problem setup from \S2 of the published paper alongside an abridged discussion and set of worked examples from \S3.
}

\begin{figure*}[t]
    \centering
    \includegraphics[width=\textwidth]{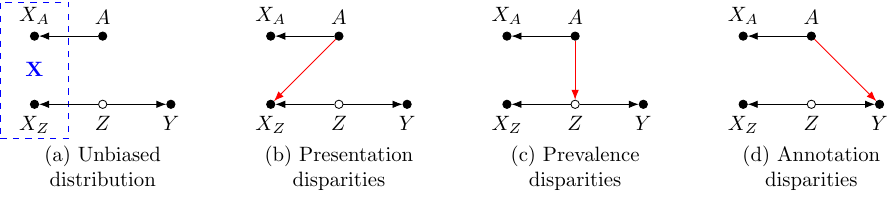}
    \caption[Causal structures of dataset bias in classification tasks.]{Causal structures of dataset bias in classification tasks. The input ${X}$ is decomposed into latent features ${X_Z, X_A}$ based on their causal relationships with the sensitive attribute $A$ and (unobserved) underlying class $Z$. In the unbiased setting (a), sensitive information is irrelevant to predicting the target $Y$. This condition may be violated by (b) feature entanglement of $A$ and $Z$, (c) differences in base rates across subgroups, or (d) differences in labelling policy across subgroups.}
    \label{fig:biasmechanisms}
\end{figure*}

Consider a medical setting where we wish to learn a fair disease classifier, given a potentially biased training dataset of images and targets. Where much work focuses on algorithmic bias \citep{chenAlgorithmicFairnessArtificial2023}, concerning errors introduced by imperfect models, we take a step back and focus on dataset bias. From a machine learning perspective, dataset bias would be reflected even by a `perfect' model with no error on its biased training dataset (or an identically distributed validation set). By focusing on what such a model would learn from a given dataset, we can use tools from causal reasoning to study how the underlying mapping between images and targets shifts across groups and settings. This is especially relevant in today's paradigm of high-capacity deep learning models, which often achieve near-zero training error when trained with the conventinoal approach of empirical risk minimisation \citep{zhangUnderstandingDeepLearning2016}.

Say we have access to a training dataset of inputs $X$, targets $Y$, and sensitive attributes $A$. The targets are a potentially noisy reflection of some unobserved underlying condition $Z$. Taking a causal interpretation, let $\mathfrak{C}_{\mathrm{tr}}$ be a structural causal model (\scm) representing the generative processes in the training dataset. Similarly, $\mathfrak{C}_{\mathrm{te}}$ is the \scm of the test dataset on which we want to make predictions. We focus on the task of learning a probabilistic model approximating the conditional test distribution $p^{\mathfrak{C}_{\mathrm{te}}}(y \mid
x)$.

The defining theme of fair image analysis is that images contain sensitive information about individuals, which models may learn to exploit inappropriately. It is thus helpful for fairness analysis to construct our causal diagram with the image decomposed into two causal factors: $X_Z$, representing the pathological structures directly caused by the disease (${Z \rightarrow X_Z}$), and $X_A$, features that encode subgroup-related sensitive information (${A \rightarrow X_A}$). Using this problem setup, we define a `no-bias' criterion (\cref{def:unbiaseddata}) and may apply the d-separation criterion \citep{vermaCausalNetworksSemantics1990} to derive three fundamental mechanisms of dataset bias (\cref{fig:biasmechanisms}).

\begin{restatable}[Unbiased distribution]{definition}{unbiaseddef} \label{def:unbiaseddata}
    The distribution induced by a structural causal model $\mathfrak{C}$ is unbiased if, given $X_Z$, sensitive information $X_A$ provides no information relevant to predicting $Y$:
\begin{align}
  Y \perp^{\mathfrak{C}} X_A \mid X_Z
  \nonumber \iff \ p^{\mathfrak{C}}(y \mid x_Z) = p^{\mathfrak{C}}(y \mid x_Z, x_A), \quad \forall(y, x_Z, x_A).
\end{align}
\end{restatable}

\subsection{Unifying fairness and distribution shift}

A core challenge in fairness analysis is defining the problem itself. One prominent approach from the literature focuses on maximising subgroup performance for test data that are independent and identically distributed (\iid) to training data, effectively ignoring dataset bias and treating fairness as a learning problem \citep[e.g.][]{zietlowLevelingComputerVision2022}. A second approach assumes that training data includes known spurious correlations and seeks to generalise to an out-of-distribution test set \citep[e.g.][]{kimLearningNotLearn2019}. A third approach even ignores absolute performance entirely, aiming instead to enforce relative equality of properties such as predicted positive \citep{zemelLearningFairRepresentations2013} or true positive \citep{hardtEqualityOpportunitySupervised2016} rates. Unfortunately, these approaches are often incompatible as they make different implicit assumptions about the nature of the bias.

Importantly, our causal setup allows us to bypass these inconsistencies by converting any fairness problem into a distribution shift problem. After specifying a causal diagram (or selecting an appropriate one from \cref{fig:biasmechanisms}), a domain expert may analyse each causal pathway to determine whether it is fair -- and information from it should be preserved -- or unfair -- and information from it should be mitigated. The fairest model in this context is one with the best possible classification performance when deployed on a (potentially hypothetical) dataset with all unfair causal pathways removed.

\begin{marginfigure}
\centering
    \includegraphics[width=\linewidth]{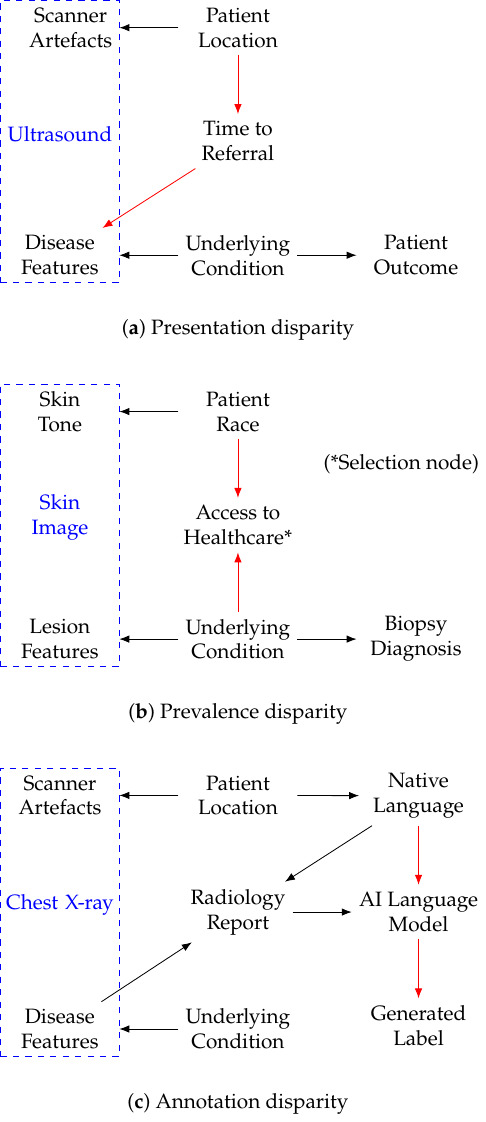}
    \caption{Worked medical examples of each causal bias mechanism. Red arrows represent unfair causal pathways. In our presentation disparity example (\textbf{a}), patients in different locations are scanned at different stages of their disease progression due to inconsistent ultrasound referral policies. In our prevalence disparity example (\textbf{b}), patient race is spuriously correlated with disease prevalence due to disparities in access to healthcare. In our annotation disparity example (\textbf{c}), disease labels are generated automatically from radiology reports with a language model and reports in different languages receive different annotation quality.}
    \label{fig:workedexamples}
\end{marginfigure}

\subsection{No fair lunch}

When viewing a fair model as one which generalises from a biased training dataset to an unbiased deployment one, a simple problem reveals itself: no model may attain perfect performance over all possible deployment settings \citep{wolpertNoFreeLunch1997}. In general, for any given training dataset, we could always construct opposing causal models which are equally compatible with the observed data \citep{hollandStatisticsCausalInference1986} and would render the same model or metric appropriate or inappropriate. This is the crux of this section. \textit{All methods for mitigating bias and metrics for measuring it must make causal assumptions about the structure of the observed dataset, including ethical assumptions about which causal pathways must be preserved or mitigated.}

In application areas where datasets have a common causal structure and unambiguous sources of bias, this issue may not be insurmountable; it may be possible to develop a standardised toolbox of methods and metrics for mitigating bias. In medical imaging, however, we deal with a wide range of dataset characteristics stemming from the use of various imaging modalities, different patient populations, clinical tasks, diagnostic processes and workflows, each contributing to the underlying causal processes with different potential sources of bias \citep{bernhardtPotentialSourcesDataset2022}. We illustrate these complications with three clinically inspired worked examples of dataset biases in \cref{fig:workedexamples}. Importantly, each of the different mechanisms may require different mitigation strategies.

\newpage
\section{The role of subgroup separability} \label{sec:sub-sep}

\marginnote[4em]{
This section is adapted from \citet{jonesRoleSubgroupSeparability2023a} (\textit{MICCAI}). We include an abridged version of the theoretical insight from \S3 of the published paper and two of the key empirical results from \S4.
}

\begin{table*}[t] 
\caption{Separability of protected subgroups in real-world datasets, measured by test-set \auc of classifiers trained to predict the groups. Mean and standard deviation are reported over ten random seeds, with results sorted by ascending mean \auc.}\label{tab:sub-sep:subgroup_sep}

\ra{1.0}
\centering
\begin{tabular}{@{}lllllcrlccc@{}}
\toprule
\textbf{Dataset-Attribute} & \phantom{ab} & \textbf{Modality} & \phantom{ab} & \multicolumn{3}{c}{\textbf{Subgroups}} & \phantom{ab} & \multicolumn{3}{c}{\textbf{AUC}} \\ 
\cmidrule(r){5-7} \cmidrule(r){9-11}
 & & & & Group 0 & & Group 1 & & $\mu$ & \phantom{a} & $\sigma$\\\midrule
 PAPILA-Sex & & Fundus Image & & Male & & Female  & & 0.642 & & 0.057 \\
 HAM10000-Sex & & Skin Dermatology & & Male & & Female  & & 0.723 & & 0.015 \\
 HAM10000-Age & & Skin Dermatology & & $<60$ & & $\geq 60$  & & 0.803 & & 0.020 \\ 
 PAPILA-Age & & Fundus Image & & $<60$ & & $\geq 60$  & & 0.812 & & 0.046 \\ 
 Fitzpatrick17k-Skin & & Skin Dermatology & & I-III & & IV-VI & & 0.891 & & 0.010 \\ 
 CheXpert-Age & & Chest X-ray & & $<60$ & & $\geq 60$  & & 0.920 & & 0.003 \\ 
 MIMIC-Age & & Chest X-ray & & $<60$ & &  $\geq 60$  & & 0.930 & & 0.002 \\ 
 CheXpert-Race & & Chest X-ray & & White & &  Non-White  & & 0.936 & & 0.005 \\
 MIMIC-Race & & Chest X-ray & & White & &  Non-White  & & 0.951 & & 0.004 \\
 CheXpert-Sex & & Chest X-ray & & Male & &  Female  & & 0.980 & & 0.020 \\ 
 MIMIC-Sex & & Chest X-ray & & Male & & Female  & & 0.986 & & 0.008 \\
 \bottomrule
 \vspace{0pt}
\end{tabular}
\end{table*}

One overlooked property of medical data is \emph{subgroup separability}: the ease with which individuals can be identified as subgroup members. Some medical images encode sensitive information that models may use to classify individuals into subgroups \citep{gichoyaAIRecognitionPatient2022}. However, this property is unlikely to hold for all modalities and protected characteristics. A more realistic premise is that subgroup separability varies across characteristics and modalities.

Our notion of subgroup separability comes from the simple insight that, in a disease prediction task, we can decompose the image to label mapping into a linear combination of the subgroup-wise mappings using the law of total probability:
\begin{equation}
    p(y \mid x) = \sum_{a \in \mathbb{S}(A)} p(y \mid x, a) p(a \mid x).
\end{equation}

\marginnote{We can think of subgroup separability as studying the $A \rightarrow X_A$ causal pathway from our setup in \cref{sec:no-fair}.}

We refer to the $p(a \mid x)$ term as \textit{subgroup separability} -- that is, it represents how much sensitive information the image encodes to begin with. When subgroup separability is high, a trained model $p_\theta$ is free to learn and use separate mappings for each subgroup
\begin{align}
    p_\theta(y \mid x) &\approx \sum_{a \in \mathbb{S}(A)} p(y \mid x, a) \mathbb{1}_{\{a'\}} (a) \\
    &\approx p(y \mid x, a'), \quad \text{where $a'$ is the true group.}
\end{align}

In contrast, when separability is low, $p(a \mid x)$ collapses to the subgroup prevalence prior $p(a)$. In this setting, models are unable to leverage subgroup information and thus must give all groups the same weighted combination of the subgroup-wise mappings.

\subsection{Subgroup separability in the real world}

We empirically test the premise of this section: subgroup separability varies across medical imaging settings. In \cref{tab:sub-sep:subgroup_sep}, we measure subgroup separability by training binary subgroup classifiers on eleven dataset-attribute combinations adapted from \citep{zongMEDFAIRBenchmarkingFairness2023}. We use test-time area under the receiver operating characteristic curve (\auc) as a proxy for separability.

Some patterns are immediately noticeable from \cref{tab:sub-sep:subgroup_sep}. All attributes can be predicted from chest X-ray scans with $> 0.9$ \auc, implying that the modality encodes substantial information about patient identity. Age is consistently well predicted across all modalities, whereas separability of biological sex varies, with prediction of sex from fundus images being especially weak. Importantly, the wide range of \auc results $[0.642 \rightarrow 0.986]$ across the dataset-attribute combinations confirms our premise that subgroup separability varies substantially across medical imaging applications. 

\subsection{Performance degradation under label bias}

\begin{figure}[t]
    \centering
    \includegraphics[width=\textwidth]{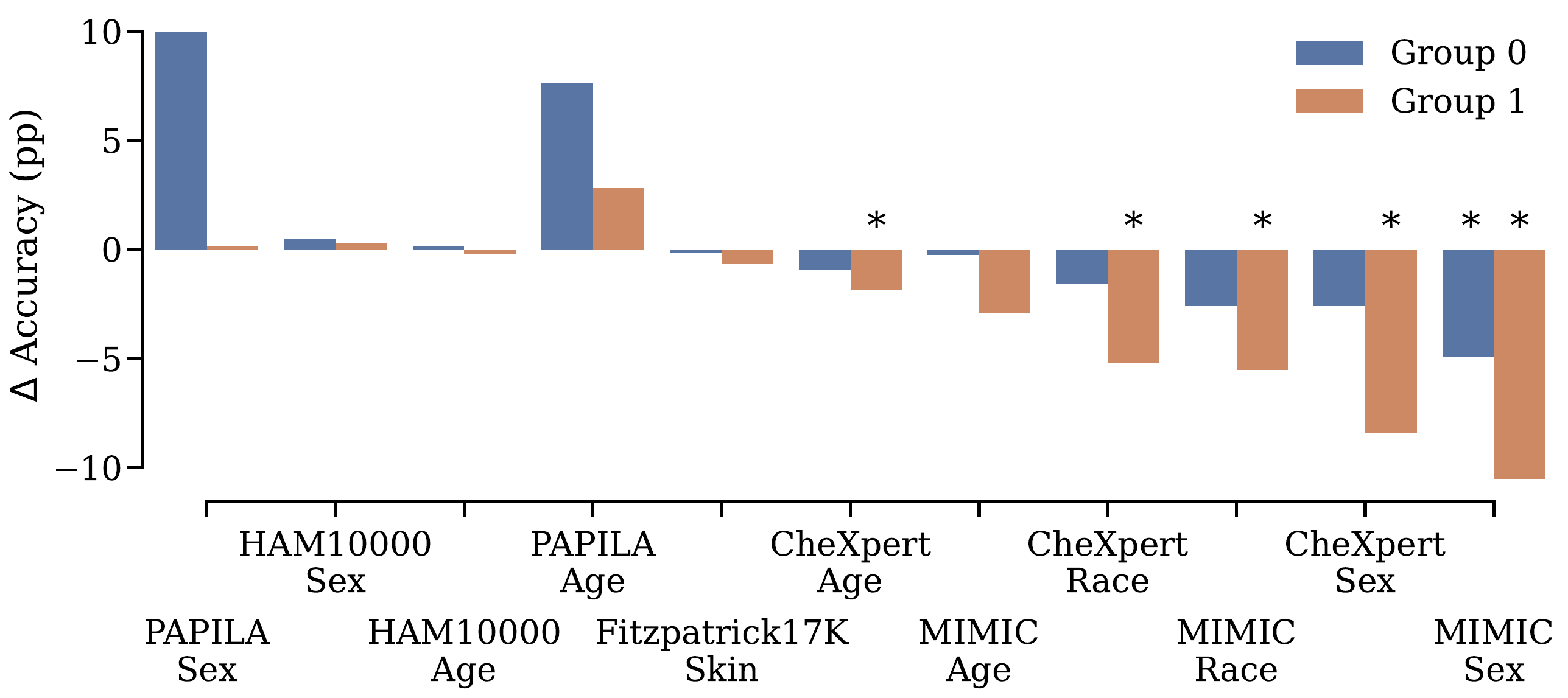}
    \caption{Percentage-point degradation in accuracy for disease classifiers trained on biased data, compared to training on clean data. Lower values indicate worse performance for the biased model when tested on a clean dataset. Results are reported over ten random seeds, and bars marked with $^*$ represent statistically significant results. Dataset-attribute combinations are sorted by ascending subgroup separability.}
    \label{fig:sub-sep:erm_degradation}
\end{figure}

We now test our theoretical insight: models are affected by underdiagnosis differently depending on subgroup separability. We inject underdiagnosis bias into each training dataset by randomly mislabelling $25 \%$ of positive individuals in Group 1 (see \cref{tab:sub-sep:subgroup_sep}) as negative. For each dataset-attribute combination, we train ten disease classification models with the biased training data and ten models with the original clean labels; we test all models on clean data. We assess how the test-time performance of the models trained on biased data degrades relative to models trained on clean data. We illustrate the mean percentage point accuracy degradation for each group in \cref{fig:sub-sep:erm_degradation} and use the Mann-Whitney U test (with the Holm-Bonferroni adjustment for multiple hypothesis testing) to determine if the performance degradation is statistically significant.

Our results in \cref{fig:sub-sep:erm_degradation} are consistent with our analysis. We report no statistically significant performance degradation for dataset-attribute combinations with low subgroup separability ($< 0.9 $ \auc). In these experiments, the proportion of mislabelled images is small relative to the total population; thus, the underdiagnosed subgroups mostly recover from label bias by sharing the correct mapping with the uncorrupted group. As subgroup separability increases, performance degrades more for the underdiagnosed group (Group 1), whilst performance for the uncorrupted group (Group 0) remains somewhat unharmed. We see a statistically significant performance drop for Group 0 in the MIMIC-Sex experiment -- we believe this is because the model learns separate group-wise mappings, shrinking the effective size of the dataset for Group 0. 

\section{Rethinking fair representations} \label{sec:rethinking-frl}
\marginnote[0em]{
This section is adapted from \citet{jones2024rethinking} (\textit{ICLR}). We include a sketch of the main theoretical result from \S4.1 of the published paper and one of the empirical results from \S5.2. We omit proofs in this abridged version.
}

In recent years, many of the most prominent methods for bias mitigation fall under the category of \textit{fair representation learning} (\frl) methods, which aim to remove sensitive information from learned representations. Motivated by our causal formulation of dataset bias, we take a detailed look at the limits of fair representations. Let's begin by recalling from \citet{zemelLearningFairRepresentations2013} that the stated aim of fair representation learning is to 
\begin{displayquote}
\textit{``lose any information that can identify whether the person belongs to
the protected subgroup, while retaining as much other
information as possible''}. 
\end{displayquote}
We will refer to the first part of this goal as \textit{effectiveness} -- is \frl effective at removing sensitive information that would have been encoded by \erm? The second part will be \textit{harmlessness} -- does \frl avoid harming performance by retaining task-relevant information? We begin by proving that fair representations cannot be both effective and harmless if test data is \iid to training data (\cref{fig:p1matrix}). 

Notably, our results follow from our causal setup in \cref{sec:no-fair}, showing how a causal approach helps to clarify complex issues in bias and fairness. We do not presuppose any architecture or implementation for the classifiers. Nor do we make assumptions about the functional mechanisms in the underlying \scm. We scrutinise the objective of learning fair representations through the lens of implied conditional independence relationships.

\begin{marginfigure}
    \centering
    \scalebox{0.59}{
    \renewcommand\arraystretch{1.5}
    \setlength\tabcolsep{0pt}
    \begin{tabular}{c >{\hspace{8pt}}r @{\hspace{0.7em}}c @{\hspace{0em}}c @{\hspace{0.7em}}l}
      \multirow{15}{*}{\rotatebox[origin=c]{90}{\bfseries \large Test}} & 
        & \multicolumn{2}{c}{\bfseries \hspace{-.7em} \large Train} & \\
      & & \large \textsc{unbiased} & \large \textsc{biased} & \\
      & \rotatebox{90}{\hspace{-2em}\large \textsc{unbiased}} & \MyBox{\textbf{Effectiveness}}{is impossible\\ (\cref{lemma:effectiveness})}{lyellow} & \MyBox{\textbf{Effectiveness} \\ \& \textbf{Harmlessness}}{are possible\\ (\cref{theorem:iidfutility})}{lgreen} & \\[4.95em]
      & \rotatebox{90}{\hspace{-1.6em} \large \textsc{biased}} & \MyBox{\textbf{Effectiveness}}{\& \textbf{Harmlessness} \\ are impossible \\ (\cref{lemma:effectiveness} \\ \& \cref{lemma:harmlessness})}{lred} & \MyBox{\textbf{Harmlessness}}{is impossible \\ (\cref{lemma:harmlessness})}{lyellow} & \\
    \end{tabular}
    }
    \vspace{1em}
    \caption{By defining concepts of \textit{Effectiveness} and \textit{Harmlessness}, we prove fundamental limitations on fair representation learning when evaluation data is identically distributed to training data.}
    \label{fig:p1matrix}
\end{marginfigure}

\subsection{Futility in \iid settings}

We consider models of the following form: a feature extractor $f_\theta$ mapping inputs to representations $R$, and a classifier which maps representations to predictions. Both components are typically implemented as (deep) neural networks. Fair representation learning imposes the train-time constraint that fair representations $R^\mathrm{FRL}$ must be (marginally) independent of the sensitive attribute, denoted as $R^\mathrm{FRL} \perp^{\mathfrak{C}_{\mathrm{tr}}} A$, leading to a predictor satisfying demographic parity. We contrast this to the unconstrained \erm strategy (i.e. learning $R^\mathrm{ERM}$). While $f_\theta$ is always a function of $X$ (i.e. the feature extractor takes the whole of $X$ as input), we will slightly abuse the notation $f_\theta(X^*)$ to indicate that the feature extractor is only non-constant w.r.t. some subset $X^*$ of $X$.

\begin{restatable}[]{assumption}{compactnessassumption} \label{assumption:compactness}
Unconstrained representations depend on inputs $X^* \subseteq X$ iff they form a Markov blanket over $Y$ at train-time:
\begin{equation} \label{eq:compactnessassumption}
    R^\mathrm{ERM} = f_\theta(X^*) \iff Y \perp^{\mathfrak{C}_{\mathrm{tr}}} (X \setminus X^*) \mid X^* .
\end{equation}
\end{restatable}
The Markov blanket contains all information sufficient to predict $Y$ in an idealised (infinite-sample) setting \citep[][Chapter 6]{petersElementsCausalInference2017}. We may view $X^*$ as a sufficient statistic for predicting $Y$; hence \cref{assumption:compactness} is closely related to the information bottleneck principle \citep{tishbyInformationBottleneckMethod2000}, which stipulates that representations should be minimal and sufficient for predicting $Y$. Intuitively speaking, \cref{assumption:compactness} states that a properly trained \erm model encodes relevant information in its representations whilst ignoring irrelevant information.

\begin{restatable}[]{lemma}{frllemma} \label{lemma:fairrepresentations}
    Fair representations must depend on $X_Z$ only: 
\begin{equation} \label{eq:fairrepresentationlemma}
    R^\mathrm{FRL} \perp^{\mathfrak{C}_{\mathrm{tr}}} A \implies R^\mathrm{FRL} = f_\theta(X_Z) .
\end{equation}
\end{restatable}

\begin{restatable}[]{lemma}{ermlemma} \label{lemma:unconstrainedrepresentations}
    Unconstrained representations are fair iff the training distribution is unbiased:
\begin{equation} \label{eq:unconstrainedrepresentationslemma}
      R^\mathrm{ERM} \perp^{\mathfrak{C}_{\mathrm{tr}}} A \iff Y \perp^{\mathfrak{C}_{\mathrm{tr}}} X_A \mid X_Z.
\end{equation}
\end{restatable}

We now take an information-theoretic perspective to define our two desiderata for fair representations: effectiveness (\cref{def:effectiveness}), and harmlessness (\cref{def:harmlessness}). While both properties are intuitive and desirable, we show how they each imply constraints on the training and testing distributions in \cref{lemma:effectiveness,lemma:harmlessness}, respectively. By showing that these constraints are incompatible when the distributions coincide, we derive our futility result for \iid settings (\cref{theorem:iidfutility}).

\begin{restatable}[Effectiveness]{definition}{effectivenessdef} \label{def:effectiveness}
    Fair representations are effective if, at train-time, they do not encode sensitive information that unconstrained representations would encode:
\begin{equation}\label{eq:effectiveness}
   I^{\mathfrak{C}_{\mathrm{tr}}}(A ; R^{\mathrm{ERM}}) > I^{\mathfrak{C}_{\mathrm{tr}}}(A ; R^\mathrm{FRL}) = 0  .
\end{equation} 
\end{restatable}
\begin{restatable}[Harmlessness]{definition}{harmlessnessdef} \label{def:harmlessness}
    Fair representations are harmless if, at test-time, they have equal information relevant to predicting the targets as the input (i.e. they do not discard relevant information).
\begin{equation} \label{eq:harmlessness}
   I^{\mathfrak{C}_{\mathrm{te}}}(Y ; R^\mathrm{FRL}) = I^{\mathfrak{C}_{\mathrm{te}}}(Y ; X_Z, X_A ) .
\end{equation}
\end{restatable}

\begin{restatable}[]{lemma}{effectivenesslemma} \label{lemma:effectiveness} Effectiveness ($\mathcal{E}$) implies bias at train-time:
\begin{equation}
    \mathcal{E} \implies Y \not\perp^{\mathfrak{C}_{\mathrm{tr}}} X_A \mid X_Z .
\end{equation}
\end{restatable}

\begin{restatable}[]{lemma}{harmlessnesslemma} \label{lemma:harmlessness} Harmlessness ($\mathcal{H}$) implies no bias at test-time:
\begin{equation}
    \mathcal{H} \implies Y \perp^{\mathfrak{C}_{\mathrm{te}}} X_A \mid X_Z .
\end{equation}
\end{restatable}

\begin{restatable}[Futility]{proposition}{futilitythm} \label{theorem:iidfutility}
    Fair representations may not be effective ($\mathcal{E}$) and harmless ($\mathcal{H}$) if the train and test datasets are identically distributed:
    \begin{equation}
        \mathcal{E} \land \mathcal{H} \implies P^{\mathfrak{C}_{\mathrm{tr}}} \not = P^{\mathfrak{C}_{\mathrm{te}}} .
    \end{equation}
\end{restatable}

We emphasise the importance of \cref{theorem:iidfutility}, given that performance-oriented \iid benchmarks persist in the literature. \textit{Fair representation learning is futile for performance-sensitive \iid tasks.} The strategy carries an implicit assumption that training data contains bias not present at test time. Intuitively, preventing a model from using information can only worsen performance unless the predictive power given by that information is entirely spurious and expected to disappear at test time. \cref{theorem:iidfutility} hence provides a theoretical explanation on why previous empirical studies benchmarking \frl methods in \iid settings did not find any consistent improvements over \erm methods~\citep{pfohlEmpiricalCharacterizationFair2021,zhangImprovingFairnessChest2022,zietlowLevelingComputerVision2022,zongMEDFAIRBenchmarkingFairness2023}. 

\subsection{Potential validity under distribution shift}

Outside the \iid setting, proving the validity of \frl is more challenging, as it would involve proving causal identifiability \citep{khemakhemVariationalAutoencodersNonlinear2020} of the $X_Z$ feature, which is not possible without reasonably strong assumptions \citep{hyvarinen2024identifiability}. Given these challenges with theoretical analysis, we focus instead on a simpler and weaker concept of validity: does \frl practically attain better performance than \erm? We propose two hypotheses, which we will explore empirically.

\begin{restatable}[]{hypothesis}{hypothesisbias} \label{hypothesis:bias}
    \frl validity under distribution shift depends on the underlying causal structure of the bias present at train-time.
\end{restatable}

\begin{restatable}[]{hypothesis}{hypothesissep} \label{hypothesis:sep}
    \frl validity under distribution shift depends on subgroup separability.
\end{restatable}

\begin{figure*}[t]
    \centering
    \includegraphics[width=\linewidth]{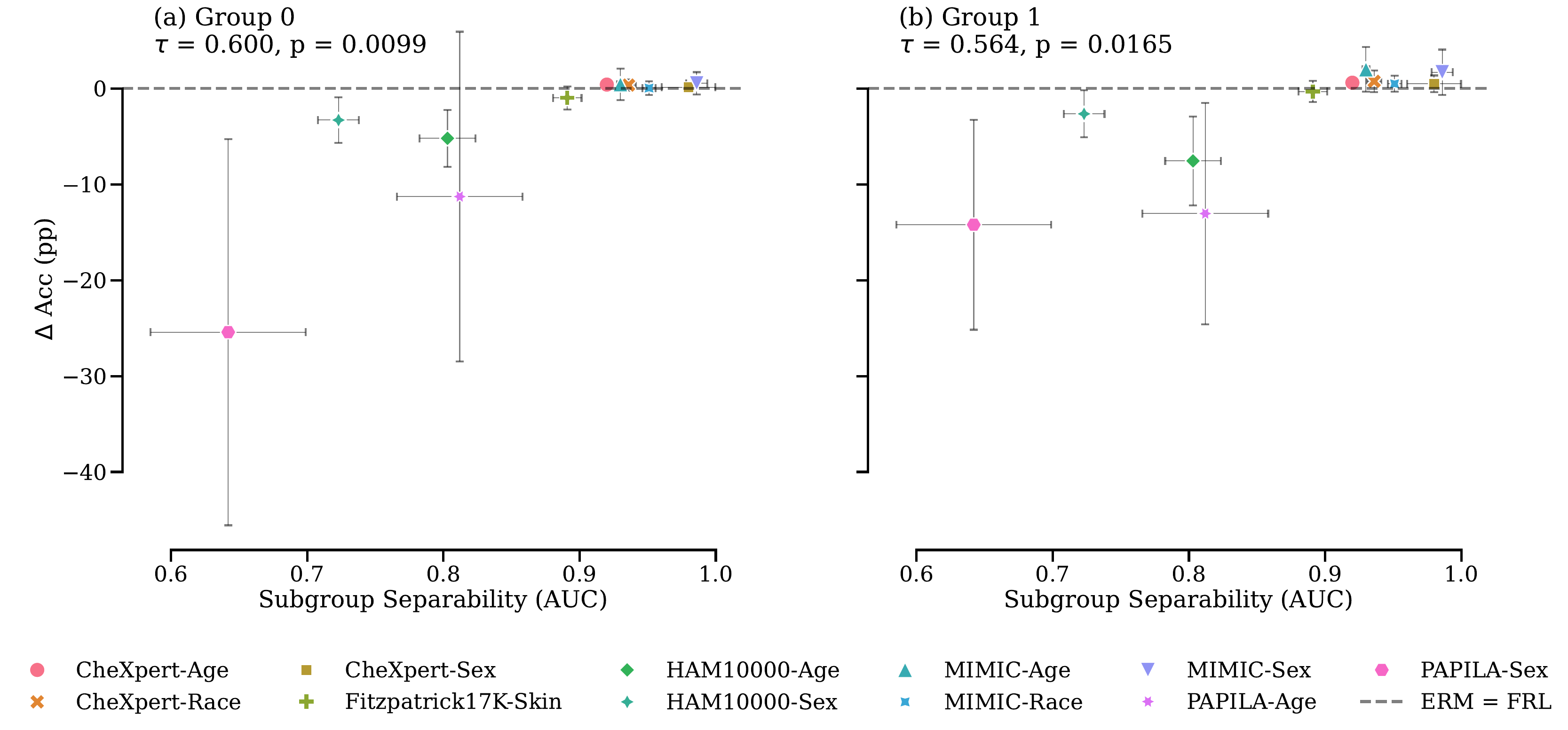}
    \caption{Percentage-point mean accuracy gap for \frl models compared to \erm models, aggregated over all bias mechanisms and plotted against subgroup separability AUC, as reported by \citet{jonesRoleSubgroupSeparability2023a}. Positive $\Delta$ Acc indicates that \frl outperforms \erm on the unbiased test set. We use Kendall's $\tau$ statistic to test for a monotonic association between $\Delta$ Acc and subgroup separability. $y$-axis error bars represent standard deviations of the aggregated $\Delta$ Acc measurements. $x$-axis error bars represent standard deviations in subgroup separability measurements.}
    \label{fig:shiftsephypothesis}
\end{figure*}

\cref{fig:shiftsephypothesis} plots the performance gap between \frl and \erm models against subgroup separability. We aggregate results for each method over nine runs (three random seeds $\times$ three bias mechanisms from \cref{fig:biasmechanisms}) and repeat for each dataset-attribute combination in from our experiments in \cref{sec:sub-sep}.

We find a statistically significant monotonic association between $\Delta$ Acc and subgroup separability, providing substantial evidence for Hypothesis \ref{hypothesis:sep}. On dataset-attribute combinations with high subgroup separability, \frl improves performance relative to \erm for the disadvantaged group (Group 1) whilst maintaining performance for the other groups. In settings with low separability, \frl substantially worsens performance for both groups. We additionally find a large variation across bias mechanisms, providing initial evidence for Hypothesis \ref{hypothesis:bias}.

\section{Outlook}

While the study of algorithmic bias is important and has gained significant interest in recent years, underlying dataset biases remain poorly understood. In \cref{sec:no-fair}, we introduced the \textit{no fair lunch} principle, demonstrating how the causal nature of dataset bias has profound consequences for deep learning algorithms. In \cref{sec:sub-sep}, we introduced \textit{subgroup separability}, a new statistical dimension to the problem based on the amount of sensitive information encoded in input images. In \cref{sec:rethinking-frl}, we applied both of these concepts to highlight the limitations of one of the most popular and widely used classes of methods in fair machine learning.

At first glance, it may appear as if these projects paint a bleak outlook for the field. However, we argue that the opposite is true. By uncovering some of the core issues harming the validity of real-world machine learning -- and by understanding why naive methods fail to solve them -- we leave the community in a stronger position to make progress on the problems.

One of the core themes of our work is that it is better to make assumptions explicit and falsifiable than to deploy methods which silently fail due to implicit or poorly understood assumptions. Causal diagrams provide a clear, mathematically principled way of making assumptions about dataset bias explicit, and we believe that approaches based on graphical modelling and causality present promising directions forward for the field.

\subsection{Acknowledgements}

We thank our coauthors on the projects presented in this primer -- as well as colleagues both within and outside Imperial College London -- for support and input. C.J. is supported by Microsoft Research, EPSRC, and The Alan Turing Institute through a Microsoft PhD scholarship and a Turing PhD enrichment award. B.G. received support from the EPSRC and the Royal Academy of Engineering through his Kheiron/RAEng Research Chair and the CHAI - EPSRC Causality in Healthcare AI Hub.

\newpage
\printbibliography
\end{document}